\documentclass[nohyperref]{article}

\usepackage{microtype}
\usepackage{graphicx}
\usepackage{subfigure}
\usepackage{booktabs} % for professional tables
\usepackage{tikz}

\usepackage{hyperref}
\usepackage{nameref}

\usepackage[accepted]{icml2022}

\usepackage{amsmath}
\usepackage{amssymb}
\usepackage{mathtools}
\usepackage{amsthm}
\usepackage{nccmath}
\usepackage{bbm}
\usepackage[shortlabels]{enumitem}
\usepackage{mdframed}
\usepackage{float}

\usepackage[capitalize,noabbrev]{cleveref}

\theoremstyle{plain}
\newtheorem{theorem}{Theorem}[section]
\newmdtheoremenv{boxedtheorem}{Theorem}[section]

\newtheorem{corollary}[theorem]{Corollary}
\theoremstyle{definition}

\newtheorem{assumption}[theorem]{Assumption}
\theoremstyle{remark}

\newcommand\E[2]{\mathbb{E}_{#1} \! \left[ #2 \right]}

\renewcommand\S{\mathbb{S}}
\newcommand\R{\mathbb{R}}
\newcommand\N{\mathcal{N}}
\renewcommand\O{\mathcal{O}}
\newcommand\relu{\textrm{ReLU}}
\newcommand\xb{\mathbf{x}}
\newcommand\zb{\mathbf{z}}

\newcommand\D{\mathcal{\D}}
\newcommand\X{\mathcal{\X}}
\newcommand\Y{\mathcal{\Y}}

\usepackage{xcolor}
\definecolor{grey}{RGB}{100,100,100}

\newcommand\numberlessfootnote[1]{%
  \begingroup
  \renewcommand\thefootnote{}\footnote{#1}%
  \addtocounter{footnote}{-1}%
  \endgroup
}

\usepackage[textsize=tiny]{todonotes}

\icmltitlerunning{Reverse Engineering the Neural Tangent Kernel}

\begin{document}

\twocolumn[
\icmltitle{Reverse Engineering the Neural Tangent Kernel}




\begin{icmlauthorlist}
\icmlauthor{James B. Simon}{phys,neuro} \ \ \
\icmlauthor{Sajant Anand}{phys} \ \ \
\icmlauthor{Michael R. DeWeese}{phys,neuro}
\end{icmlauthorlist}

\icmlaffiliation{phys}{Department of Physics, University of California, Berkeley, Berkeley, CA 94720}
\icmlaffiliation{neuro}{Redwood Center for Theoretical Neuroscience and Helen Wills Neuroscience Institute, University of California, Berkeley, Berkeley, CA 94720}

\icmlcorrespondingauthor{James B. Simon}{james.simon@berkeley.edu}

\icmlkeywords{deep learning, deep learning theory, NNGP, NTK, neural tangent kernel}

\vskip 0.3in
]



\printAffiliationsAndNotice  % leave blank if no need to mention equal contribution

\begin{abstract}
The development of methods to guide the design of neural networks is an important open challenge for deep learning theory.
As a paradigm for principled neural architecture design, we propose the translation of high-performing \textit{kernels}, which are better-understood and amenable to first-principles design, into equivalent \textit{network architectures}, which have superior efficiency, flexibility, and feature learning.
% using recent network-kernel correspondences.
To this end, we constructively prove that, with just an appropriate choice of activation function, any positive-semidefinite dot-product kernel can be realized as either the NNGP or neural tangent kernel of a fully-connected neural network \textit{with only one hidden layer}.
We verify our construction numerically and demonstrate its utility as a design tool for finite fully-connected networks in several experiments.
\numberlessfootnote{\\Code to reproduce experiments is available at \url{https://github.com/james-simon/reverse-engineering}.}
\end{abstract}

\section{Introduction}
\label{sec:intro}

The field of deep learning theory has recently been transformed by the discovery that, as network widths approach infinity, models take simple analytical forms amenable to theoretical analysis. These limiting forms are described by kernel functions, the most important of which is the \textit{neural tangent kernel} (NTK), which describes the training of an infinite-width model via gradient descent \citep{jacot:2018, lee:2019-ntk}.
Limiting kernels are now known for fully-connected networks \citep{daniely:2016, lee:2018-nngp, matthews:2018-nngp, jacot:2018, lee:2019-ntk}, convolutional networks \citep{novak_cnngp:2019, arora:2019-cntk}, transformers \citep{hron:2020-transformer-ntk}, and more \citep{yang:2019-tensor-programs-I}, giving unprecedented insight into commonly-used network architectures.

However, deep learning theory must ultimately aim not merely to understand existing architectures but to \textit{design better architectures}. Neural network design is famously unscientific, proceeding via an artful combination of intuition, trial-and-error, and brute-force search that yields little understanding of why certain architectures perform well or how we might do better. We argue that theorists should endeavor to use the many insights gained from the study of wide networks to develop well-understood methods and principles to illuminate this process.

We propose that one potential avenue for principled architecture design is \textit{reverse engineering the network-kernel correspondence}: first design a kernel well-suited to the task at hand, then choose network hyperparameters to (approximately) realize that kernel as a finite network. This paradigm promises the best of both model types: kernels are analytically simple and thus far more amenable to first-principles design, while neural networks have the advantages of cheaper training/inference, flexibility in training schemes and regularization, and the ability to learn useful features and transfer to new tasks.

Despite clear implications for architecture design as well as fundamental interest, very little work has studied the problem of achieving a desired kernel with a network architecture. \textit{Here we fully solve this problem for fully-connected networks (FCNs) on normalized data:} we specify the full set of achievable FCN kernels and provide a construction to realize any achievable NNGP kernel or NTK with a shallow (i.e., single-hidden-layer) FCN. We then confirm our results experimentally and describe two cases in which, by reverse-engineering a desired NTK, we can design high-performing FCNs in a principled way.

Our main contributions are as follows:
\begin{itemize}
    \item We constructively prove that, with just an appropriate choice of activation function, any positive-semidefinite dot-product kernel can be realized as the NNGP kernel or NTK of an FCN with no biases and just one hidden layer.
    \item As a surprising corollary, we prove that shallow FCNs can realize a strictly \textit{larger} set of dot-product kernels than FCNs with multiple nonlinear layers can.
    \item We experimentally verify our construction and demonstrate our reverse engineering paradigm for the design of FCNs in two experiments: (a) we engineer a single-hidden-layer, finite-width FCN that mimics the training and generalization behavior of a deep ReLU FCN over a range of network widths, and (b) we design an FCN that significantly outperforms ReLU FCNs on a synthetic parity problem.
\end{itemize}

\subsection{Related Work}

The connection between kernels and infinitely wide networks was first revealed by \citet{neal:1996}, who showed that randomly-initialized single-hidden-layer FCNs approach Gaussian processes with deterministic kernels as their width increases. Much later, this work was extended to randomly-initialized deep FCNs \citep{lee:2018-nngp, matthews:2018-nngp}, and the kernels were given the name ``neural network Gaussian process" (NNGP) or ``conjugate" kernels. The related NTK, describing the training of deep FCNs, was discovered soon thereafter \citep{jacot:2018}. These findings and their extensions to other architectures have enabled many insights into neural network training \citep{zou:2018-convergence, du:2019-convergence, allen:2019-convergence} and generalization \citep{arora:2019-NTK-gen-bound, bordelon:2020-learning-curves, simon:2021-eigenlearning}. Many studies have analyzed the mapping from network architectures to corresponding kernels, but very few have considered the inverse mapping, which is the focus of this work.

Anticipating the NNGP correspondence, \citet{daniely:2016} considered the kernels represented by wide randomly-initialized networks of various architectures. These deep kernels are given by the composition of many layerwise kernels, and for FCNs, they derive an invertible mapping between activation function and layerwise kernel. Many of our results for the NNGP kernel follow from their work. We note that that study did not consider training or the NTK, did not discuss achieving a desired kernel as a design principle, and included no experiments.

Our work has close connections to random-feature models \citep{rahimi:2007, rahimi:2008}, which use random feature embeddings to approximate analytical kernels. Shallow networks with frozen first layers are random feature models approximating NNGP kernels (though here we chiefly consider the NTK and train all layers in our experiments). Unlike with neural networks, there is a substantial literature on engineering random features to yield desired kernels, with feature maps known for arbitrary dot-product kernels \citep{kar:2012-rf_dot_product-kernels, pennington:2015-rf_dot_product-kernels}.

Our equivalence between deep and shallow FCNs is related to the titular observation of \citet{bietti:2020-deep-equals-shallow} that ReLU FCN NTKs of all depths have the same asymptotic eigendecay.
We extend this observation, noting that, with an informed modification of the shallow FCN's activation function, it can have not merely the same kernel eigendecay but in fact the \textit{same kernel} as its deep counterpart.

In a concurrent study in a spirit similar to ours, \citet{shi:2021} also explored NTK performance as a principle for activation function design.
Like us, they decompose activation functions in a Hermite basis, but optimize the coefficients to directly minimize NTK test MSE in a black-box fashion, whereas in our comparable parity problem experiment, we transparently choose a suitable kernel matching the symmetry of the problem.

\section{Theoretical Background}
\label{sec:th}

\subsection{Preliminaries and Notation}

We study fully connected architectures in this work. An FCN with $L$ hidden layers of widths $\{n_\ell\}_{\ell=1}^{L}$ is defined by the recurrence relations

\begin{equation}
    \zb^{(\ell)} = W^{(\ell)} \xb^{(\ell-1)} + \mathbf{b}^{(\ell)}, \ \ \ \ \ \ \ \xb^{(\ell)} = \phi(\zb^{(\ell)}),
\end{equation}

starting with the input vector $\xb^{(0)} \in \R^{d_{\textrm{in}}}$ and ending with the output vector $\zb^{(L+1)} \in \R^{d_{\textrm{out}}}$. The weight matrices and bias vectors are defined and initialized as

\small
\begin{align}
    W^{(\ell)} \!&\in\! \R^{n_{\ell} \times n_{\ell-1}}, \ \ \
    W^{(\ell)}_{ij} \!=\! \frac{\sigma_w}{\sqrt{n_\ell}} \omega^{(\ell)}_{ij},
    &\omega^{(\ell)}_{ij} \sim \N(0,1), \\
    \mathbf{b}^{(\ell)} \!&\in\! \R^{n_{\ell}}, \qquad \ \ \ \ \ \ \
    \mathbf{b}^{(\ell)}_{i} \!=\! \sigma_b \beta^{(\ell)}_{i},
    &\beta^{(\ell)}_{i} \sim \N(0,1),
\end{align}
\normalsize

where $\sigma_w$ and $\sigma_b$ define the initialization scales of the weights and biases, $\omega^{(\ell)}_{ij}$ and $\beta^{(\ell)}_{i}$ are the \textit{trainable parameters}, and $n_0 = d_{\textrm{in}}$ and $n_{L+1} = d_{\textrm{out}}$. It is the trainable parameters, not the weights and biases themselves, that will evolve according to gradient descent. This formulation, commonly called the ``NTK parameterization," effectively decreases the weight updates by a factor of $\O(n_\ell^{-1})$ and allows us to take a sensible infinite-width limit of the network dynamics. The initialization scales $\sigma_w,\sigma_b$ can in principle vary between layers, but we will generally take them to be layer-independent. We note that setting $\sigma_b=0$ corresponds to a network with no biases.

% Let $\thetab$ be the vector of all trainable parameters, and fix loss function $\mathcal{L}$, learning rate $eta$, and dataset $\D \equiv \{\xb_i\}_{i=1}^D$. Under full-batch gradient descent, $\thetab$ evolves according to
% \begin{equation}
%     \thetab \rightarrow \thetab - \eta \nabla_\thetab \L(\mathbf{y}, \hat{\mathbf{y}})
% \end{equation}

Throughout our analysis,
unless stated otherwise, we will make the following normalization assumption on the data:
\begin{assumption}
\label{assm:normalization}
Each input $\xb_i$ satisfies $|x_i| = \sqrt{n^0}$.
\end{assumption}
Though this assumption is rarely satisfied by natural data, enforcing it costs only one degree of freedom, and it can be enforced invertibly by appending one dummy index to each input before normalization so the number of degrees of freedom is preserved.
This assumption will greatly simplify our theoretical analysis of FCN kernels.
We will later explore the consequences of lifting this assumption empirically.

We will use $n!!$ to denote the double factorial, $\S^n$ to represent
the n-sphere, and $\delta_{ij}$ to mean $\mathbbm{1}_{i=j}$. We will write the first derivative of a function $f(z)$ as either $f'(z)$ or $\partial_z f(z)$. We will also abbreviate the bivariate centered normal distribution with diagonal $\sigma^2$ and correlation coefficient $c$ as
\begin{equation}
    \N^{\sigma^2}_c \equiv
    \N\left(
    \begin{bmatrix}
    0 \\
    0
    \end{bmatrix},
    \begin{bmatrix}
    \sigma^2 & c \, \sigma^2 \\
    c \, \sigma^2 & \sigma^2
    \end{bmatrix}
    \right).
\end{equation}

\subsection{Neural Network Kernels}
As a neural network's width approaches infinity, its behavior and properties become simple and tractable. In particular, its limiting behavior is described by two kernels in two different senses.\footnote{A \textit{kernel} is essentially a bounded, symmetric, positive-semidefinite scalar function of two variables. We refer unfamiliar readers to \citet{shawe:2004} for an introduction.}

\textbf{The NNGP.} A network with all its parameters randomly initialized will represent a random function, with the prior over parameters determining the prior over functions. In the infinite-width limit, this random function is sampled from a Gaussian process over the input space with mean zero and covariance given by an NNGP kernel $K^{(\text{NNGP})}(\cdot, \cdot)$ \citep{matthews:2018-nngp, lee:2018-nngp}. Conditioning this functional prior on training data corresponds to kernel (ridge) regression with the NNGP kernel.

\textbf{The NTK.} When performing gradient flow on a training sample $\xb_1$, the network output $f$ on another sample $\xb_2$ changes at a rate proportional to $K^{(\text{NTK})}(\xb_1, \xb_2) \equiv \nabla_\theta f(\xb_1) \cdot \nabla_\theta f(\xb_2)$. Remarkably, knowledge of the model's NTK at all times is sufficient to describe training and generalization dynamics, with no knowledge of the model's internal structure required. Even more remarkably, for an infinite-width network, the NTK is both initialization-independent and fixed throughout training, and in many cases this limiting NTK can be computed in closed form \citep{jacot:2018}. An ensemble of such wide networks trained for infinite time is equivalent to kernel regression with the NTK \citep{jacot:2018, lee:2019-ntk}.

When referring to the kernels of FCN architectures, we will generally mean their deterministic kernels at infinite width. We will explicitly note when, in later experiments, we examine the empirical NTKs of finite networks.

\textbf{Dot-product kernels.} For FCNs, both the limiting NNGP and limiting NTK kernels are \textit{rotation-invariant}: $K(\xb_1, \xb_2) = K(|\xb_1|, |\xb_2|, c)$, with $c \equiv \frac{\xb_1 \cdot \xb_2}{|\xb_1||\xb_2|} \in [-1,1]$. Fixing $|\xb_1|$ and $|\xb_2|$ as per Assumption \ref{assm:normalization}, only the third argument can vary, and the functional form reduces to merely $K(c)$. We call kernels of this form \textit{dot-product kernels.} We note that an FCN's kernels are independent of $d_{\textrm{in}}$.

We will make use of the following classic result governing dot-product kernels. The nonnegativity constraints result from the requirement that the kernel be positive-semidefinite.

\begin{theorem}[\citet{schoenberg:1942}]
\label{thm:psd}
Any dot product kernel over $\S^\infty \times \S^\infty$ must take the form $K(c) = \sum_{i=0}^\infty a_i c^i$, with $a_i \ge 0$ and $\sum_{i=0}^\infty a_i < \infty$.
\end{theorem}

We will say that a polynomial $K : [-1,1]\rightarrow\R$ is positive-semidefinite (PSD) if it obeys the constraints of Theorem \ref{thm:psd}. We note that the set of allowable dot-product kernels over $\S^{d_{\textrm{in}}} \times \S^{d_{\textrm{in}}}$ with finite $d_{\textrm{in}}$ is somewhat larger, with the conditions on $K$ becoming more restrictive and approaching those of Theorem \ref{thm:psd} as $d_{\textrm{in}} \rightarrow \infty$. However, since an FCN's kernels are independent of input dimension, they must obey Theorem \ref{thm:psd}, and we neglect kernels valid only at finite $d_{\textrm{in}}$.

\textbf{Kernels of 1HL networks.} Here we explicitly write the two kernels of single-hidden-layer (1HL) networks with activation function $\phi$ and initialization scales $\sigma_w, \sigma_b$ (see, e.g., Appendix E of \citet{lee:2019-ntk}). We make use of the $\tau$-transform, a common functional in the study of infinite-width networks, defined as
\begin{align} \label{eqn:tau_transform_def}
    \tau_\phi(c ; \sigma^2) &\equiv \E{z_1,z_2 \sim \N^{\sigma^2}_c}{\phi(z_1)\phi(z_2)}.
    % \dot{\tau}_\phi(c ; \sigma^2) &\equiv \E{z_1,z_2 \sim \N^{\sigma^2}_c}{\phi'(z_1)\phi'(z_2)}.
\end{align}
When the second argument to $\tau$
% or $\dot{\tau}$
is $1$, we will simply omit it and write $\tau(c)$. We note the following useful identity, which we prove in Appendix \ref{app:proofs}:
\begin{equation} \label{eqn:tau_tau_dot_identity}
    % \dot{\tau}_\phi(c; \sigma^2) = \frac{\partial_c}{\sigma^2} \tau_\phi(c; \sigma^2).
    \tau_{\phi'}(c; \sigma^2) = \frac{\partial_c}{\sigma^2} \tau_\phi(c; \sigma^2).
\end{equation}
In terms of the $\tau$-transform, the NNGP and NTK kernels of a 1HL network are
\begin{align}
    \label{eqn:1HL_nngp}
    K^{(\text{NNGP})}(c) &= \sigma_w^2 \tau_\phi\left(
    \frac{\sigma_w^2 c + \sigma_b^2}{\sigma_w^2 + \sigma_b^2} ;
    \sigma_w^2 + \sigma_b^2
    \right) + \sigma_b^2, \\
    \label{eqn:1HL_ntk}
    K^{(\text{NTK})}(c) &= K^{(\text{NNGP})}(c) \\
    &+ (\sigma_w^2 c + \sigma_b^2)
    % \dot{\tau}_\phi\left(
    \tau_{\phi'}\left(
    \frac{\sigma_w^2 c + \sigma_b^2}{\sigma_w^2 + \sigma_b^2} ;
    \sigma_w^2 + \sigma_b^2
    \right). \nonumber
\end{align}

\subsection{Hermite Polynomials}

The \textit{Hermite polynomials} $h_0, h_1, h_2, ...$ are an orthonormal basis of polynomials obtained by applying the Gram-Schmidt process to $1, z, z^2, ...$ with respect to the inner product $\langle f, g \rangle = \frac{1}{\sqrt{2\pi}} \int_{-\infty}^{\infty} e^{-z^2 / 2} f(z) g(z) dz$. As the Hermite polynomials form a complete basis, any function $\phi$ that is square-integrable w.r.t. the Gaussian measure can be uniquely decomposed as $\phi(z) = \sum_{k=0}^{\infty} b_k h_k(z)$. We defer a more detailed introduction to the Hermite polynomials to Appendix \ref{app:hermite} and here just mention two properties that will be of particular use (see e.g. \citet{odonnell:2014} and \href{https://en.wikipedia.org/wiki/Hermite_polynomials}{Wikipedia}):

\begin{align}
    \label{eqn:hermite_prop_double_int}
    \E{z_1,z_2 \sim \N^1_c}{h_k(z_1) h_\ell(z_2)} &= c^k \delta_{k\ell} \ \ \ \forall c \in [-1,1], \\
    \label{eqn:hermite_prop_derivative}
    h_k'(z) &= \sqrt{k} h_{k-1}(z) \ \ \ \forall k \ge 1.
\end{align}

\section{Theoretical Results}
We now present our main theoretical results. We note that the results for NNGP kernels follow from Lemma 11 of \citet{daniely:2016}; we include them for completeness, but the main new results are those for the NTK.

% RESULTS BOX
\noindent\fbox{
\parbox{79mm}{
\begin{theorem}[Kernel reverse engineering]
\label{thm:rev}
Any desired dot product kernel $K(c) = \sum_{k=0}^\infty a_k c^k$ can be achieved as
\vspace{-3mm}
\begin{itemize}
    \item the NNGP kernel of a single-hidden-layer FCN with $\sigma_w=1, \sigma_b=0$ and $\phi(z) = \sum_{k=0}^\infty \pm a_k^{1/2} h_k(z)$,
    \item the NTK of a single-hidden-layer FCN with $\sigma_w=1, \sigma_b=0$ and $\phi(z) = \sum_{k=0}^\infty \pm \! \left(\frac{a_k}{1+k}\right)^{1/2} h_k(z)$,
\end{itemize}
where we use $\pm$ to indicate that the sign of each term is arbitrary. These are the complete set of activation functions that give this kernel in a single-hidden-layer FCN with $\sigma_w=1, \sigma_b=0$.
\end{theorem}
\begin{corollary} \label{cor:rev_prop}
Let $\phi(z) \longleftrightarrow K(c)$ signify that $K$ is the NNGP kernel or NTK of a single-hidden-layer FCN with activation function $\phi$. If $\sigma_b=0$ and $\phi(z) \longleftrightarrow K(c)$, then
\begin{enumerate}[(a)]
    \item $\phi'(z) \longleftrightarrow \sigma_w^{-2} K'(c)$.
    \item $\alpha \phi(z) \longleftrightarrow \alpha^2 K(c)$ for all $\alpha \in \R$.
    \item $\phi(-z) \longleftrightarrow K(c)$.
\end{enumerate}
\end{corollary}
\begin{corollary} \label{cor:impotence_of_depth}
There are dot-product kernels that can be the NNGP kernel or NTK of a FCN {\normalfont only if} it has exactly one nonlinear hidden layer.
\end{corollary}
}
}

The proof of Theorem \ref{thm:rev} is remarkably simple, and we provide it here. We defer the proofs of Corollaries \ref{cor:rev_prop} and \ref{cor:impotence_of_depth} to Appendix \ref{app:proofs}.

\textbf{Proof of Theorem \ref{thm:rev}}.
First, we observe from Equations \ref{eqn:tau_tau_dot_identity}, \ref{eqn:1HL_nngp} and \ref{eqn:1HL_ntk} that a 1HL FCN with $\sigma_w=1, \sigma_b=0$ has kernels
\begin{align}
    \label{eqn:1HL_nngp_simple}
    K^{(\text{NNGP})}(c)
    &= \tau_\phi(c), \\
    \label{eqn:1HL_ntk_simple}
    K^{(\text{NTK})}(c)
    % &= \tau_\phi(c) + c \dot{\tau}_\phi(c)
    &= \tau_\phi(c) + c \tau_{\phi'}(c)
    = (1 + c \partial_c) \tau_\phi(c).
\end{align}
Next we observe that, for an activation function $\phi(z) = \sum_{k=0}^\infty b_k h_k(z)$, Equation \ref{eqn:hermite_prop_double_int} lets us evaluate the $\tau$-transforms, yielding
\begin{align}
    K^{(\text{NNGP})}(c) = \tau_\phi(c) &= \sum_{k=0}^\infty b_k^2 c^k, \\
    K^{(\text{NTK})}(c) = (1 + c \partial_c) \tau_\phi(c) &= \sum_{k=0}^\infty b_k^2 (1 + k) c^k.
\end{align}
Equating these kernels with the desired $K(c) = \sum_{k=0}^\infty a_k c^k$ and solving for $b_k$ completes the proof.
\qedsymbol

\textbf{FCNs have maximal kernel expressivity.}
The traditional notion of \textit{expressivity} denotes the set of functions achievable by varying a model's parameters.
Our results can be understood in terms of a new notion of \textit{kernel expressivity}, denoting the set of kernels achievable by varying an architecture's hyperparameters.
Theorem \ref{thm:rev} states that, for both NNGP and NTK kernels, FCNs have the greatest kernel expressivity we might have hoped for even with just one hidden layer: they can achieve all dot product kernels on $\S^\infty \times \S^\infty$ (i.e. those obeying the condition of Theorem \ref{thm:psd}).

\textbf{Depth is inessential.}
We emphasize the surprising fact that, as all achievable dot-product kernels are achievable with a single hidden layer as per Theorem \ref{thm:rev}, \textit{additional hidden layers do not increase kernel expressivity.}
In fact, Corollary \ref{cor:impotence_of_depth} states that additional nonlinear layers in fact \textit{decrease} kernel expressivity.\footnote{We emphasize that Corollary \ref{cor:impotence_of_depth} is true even permitting deep networks to use different nonlinearities at different layers. The proof of this result relies on the fact that certain PSD polynomials cannot be written as the composition of two other nonlinear PSD polynomials. We note that additional layers with \textit{linear} activations do not decrease kernel expressivity.}
These observations contradict the widespread belief that deeper networks are fundamentally more expressive and flexible (e.g. \citet{lecun:2015-deep-learning, poggio:2017, poole:2016, schoenholz:2017, telgarsky:2016, raghu:2017, mhaskar:2017, lin:2017, rolnick:2018}) and suggest that, at least for wide FCNs, one hidden layer is all you need.

\textbf{Biases are unnecessary.}
It is also surprising that the full space of achievable dot product kernels can be achieved with only weights and no biases (i.e. $\sigma_b=0$).
This suggests that biases are unimportant when the activation function is chosen well and are merely neuroscience-inspired holdouts from the early days of deep learning.
% This conclusion is supported by the fact that many modern architectures (e.g. ResNet \citep{he:2016}) typically do not use biases.

% KERNEL GRID FIGURE
\begin{figure*}
  \centering
  \includegraphics[width=16cm]{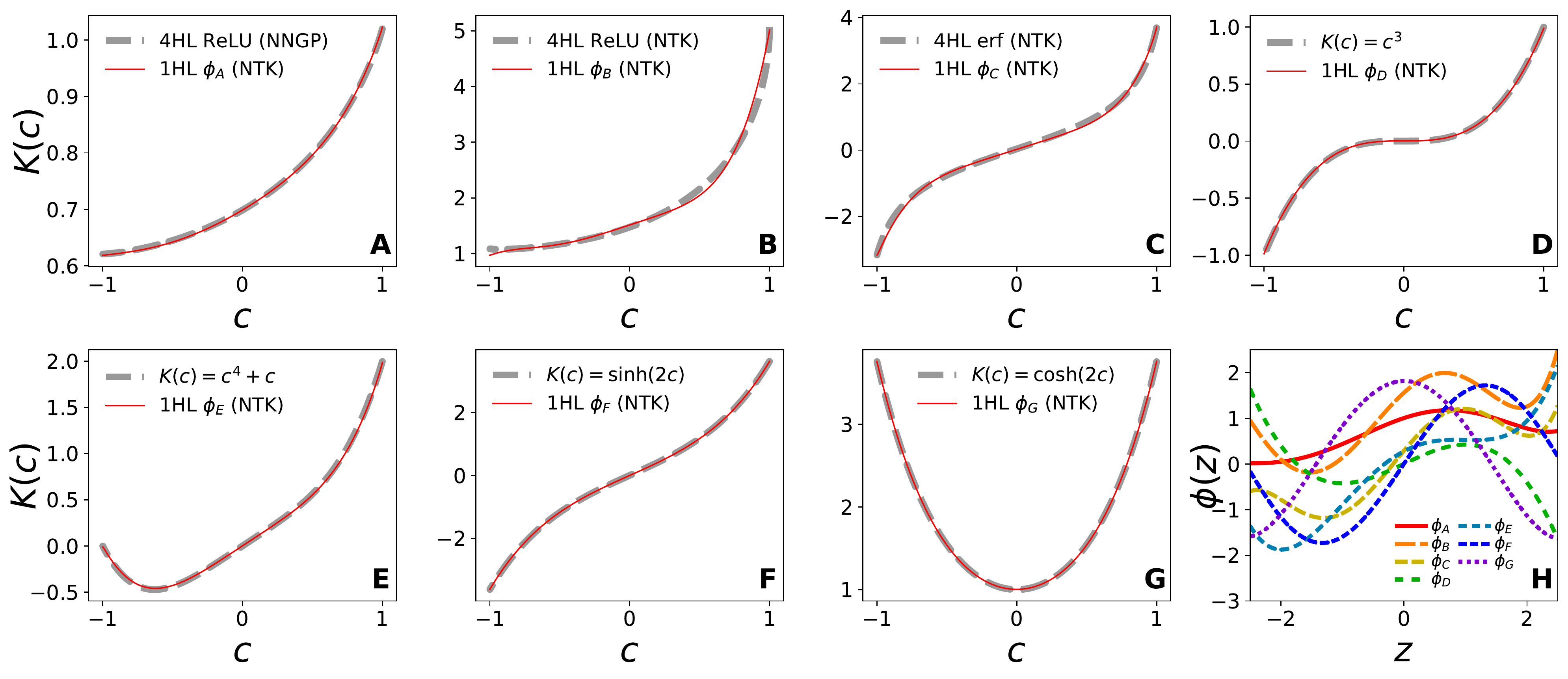}
  \caption{
  \textbf{Any dot-product kernel can be realized as the NTK of a single-hidden-layer FCN.}
  \textbf{(A-G)} Various desired kernels (grey dashed curves) and empirical NTKs of single-hidden-layer networks (solid red curves). \textbf{(H)} The engineered activation functions used in (A-G), which are all polynomials of degree at most 5.
  }
  \label{fig:kernel_grid}
\end{figure*}
 
\textbf{NTKs for selected shallow networks.}
Prior works have computed $\tau_\phi(c)$ for several choices of $\phi$; we reproduce these expressions in Appendix \ref{app:hermite}.
% With the observation that, for a 1HL network with $\sigma_w=1, \sigma_b=0$, the NTK obeys $K^{(\text{NTK})}(c) = (1 + c \, \partial_c) K^{(\text{NNGP})}(c)$, 
Applying Equation \ref{eqn:1HL_ntk_simple},
we can then easily compute the 1HL NTKs for these choices of $\phi$:

\begin{align}
    K_{e^{az}}^{(\text{NTK})} &= e^{a^2(1 + c)}(1 + a^2 c), \\
    \label{eqn:sin_kernel}
    K_{\sin(az)}^{(\text{NTK})} &= e^{-a^2} \left(\sinh(a^2 c) + a^2 c \cosh(a^2 c)\right), \\
    K_{\cos(az)}^{(\text{NTK})} &= e^{-a^2} \left(\cosh(a^2 c) + a^2 c \sinh(a^2 c)\right), \\
    K_{\relu(z)}^{(\text{NTK})} 
    &= \frac{1}{2\pi} \left( \left(1-c^2\right)^{1/2} + 2c\left(\pi - \cos^{-1}(c)\right) \right).
\end{align}

We will revisit the sinusoidal kernels experimentally in Section \ref{subsec:exp_parity}.
We note that the ReLU kernel is bounded but has divergent slope at $c \rightarrow 1$, a hallmark of ReLU  NTKs apparent at all depths.

\section{Experimental results for finite networks}
\label{sec:exp}
Our theory shows that arbitrary dot product NNGP kernels or NTKs can be achieved in infinitely-wide 1HL networks with a suitable choice of activation function.
Here we explore the experimental implications for finite networks.

All experiments use \texttt{JAX} \citep{bradbury:2018-jax} and \texttt{neural\_tangents} \citep{novak_nt:2019} for network training and kernel computation. Unless otherwise stated, all datasets are normalized to satisfy Assumption \ref{assm:normalization}.
% , but in Appendix \ref{app:additional_figs} we demonstrate that relaxing this condition such that each input $\xb_i$ \textit{on average} lies on a hypersphere of radius $\sqrt{n^0}$  is not entirely detrimental to the kernel-network mapping; .
We report full hyperparameters and provide additional commentary for each experiment in Appendix \ref{app:exp_details}.

% MIMIC KERNEL FIGURE
\begin{figure*}
  \centering
  \includegraphics[width=17cm]{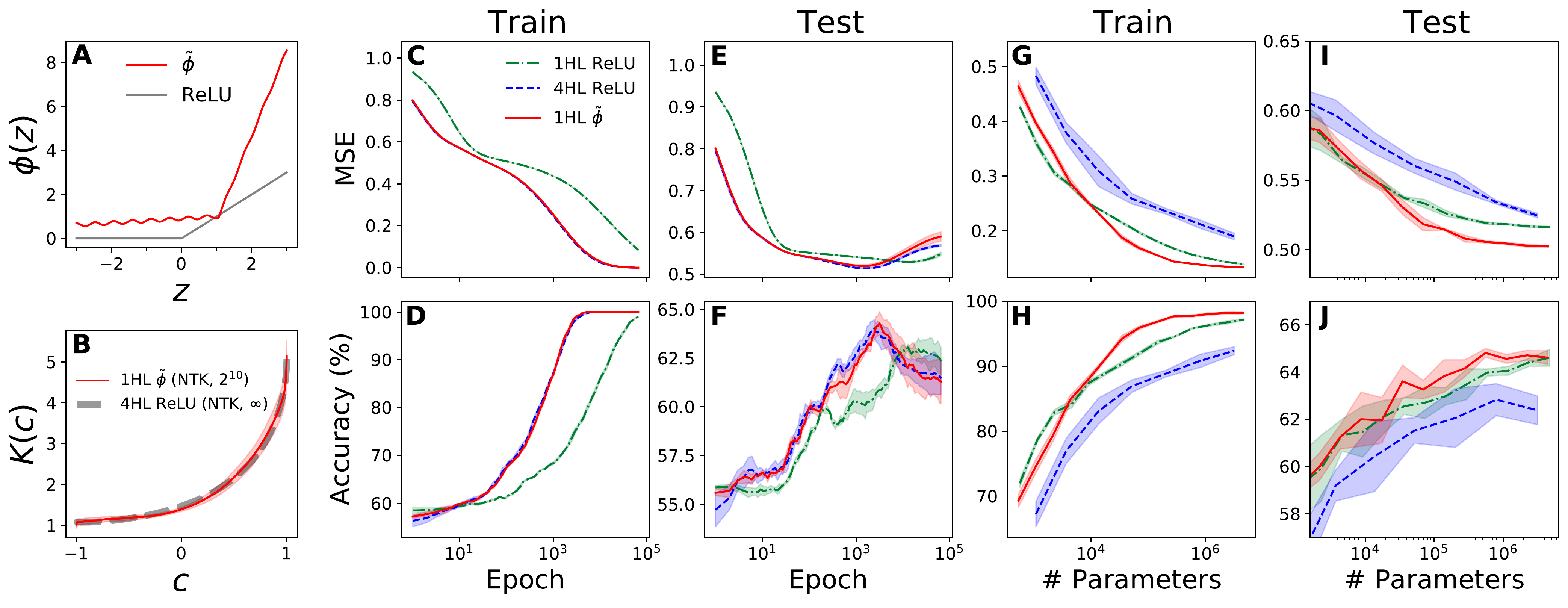}
  \caption{
  \textbf{A 1HL network with activation function designed to mimic a 4HL ReLU NTK achieves the deep architecture's training and generalization behavior with far fewer parameters.}
  \textbf{(A)} The optimized activation function $\tilde\phi$ designed to mimic the infinite-width NTK of a 4HL ReLU network, with ReLU for comparison.
  \textbf{(B)} The empirical NTK of a width-$2^{10}$ 1HL network with activation function $\tilde\phi$, with the target deep NTK for comparison.
  In all subplots, shaded regions show $1\sigma$ variation due to random initialization.
  \textbf{(C-F)} Train and test MSE and accuracy for the \texttt{wine-quality-red} task throughout training for finite 1HL ReLU, 4HL ReLU, and 1HL $\tilde\phi$ networks. Metrics for the shallow $\tilde\phi$ network closely match those of the deep ReLU network.
  \textbf{(G-J)}. Final train and test MSE and accuracy for the three architectures with various widths, plotted as a function of the total number of parameters.
  }
  \label{fig:mimic_main}
\end{figure*}

\subsection{Arbitrary NTKs with a Single Hidden Layer}
\label{subsec:exp_ntks}

To begin, we illustrate our main result by exhibiting 1HL networks that approximate desired NTKs.
For each of seven desired kernels $K(c)$, including three kernels of deep networks, we compute a 5th-order polynomial fit (i.e. approximate $K(c) \approx \sum_{k=0}^5 a_k c^k$) and use Theorem \ref{thm:rev} to construct a suitable polynomial activation function $\phi$. We then randomly initialize a 1HL network with $\sigma_w=1, \sigma_b=0$ and width $2^{18}$ and compute its empirical NTK as a function of $c$ with toy data $\left\{\left(\sqrt{2} \cos(\theta), \sqrt{2} \sin(\theta)\right)\right\}_{\theta=0}^\pi$.\footnote{The factors of $\sqrt{2}$ serve to enforce Assumption \ref{assm:normalization}.}
Using the first point in the sequence as one kernel argument, $c = \cos(\theta)$.

The results are shown in Figure \ref{fig:kernel_grid}. In each case, including when mimicking deep neural networks kernels, we find that the 1HL network's empirical NTK is remarkably close to the target kernel. The seven polynomial activation functions are plotted in Figure \ref{fig:kernel_grid}H.

Discrepancies between target and empirical kernels in Figure \ref{fig:kernel_grid} are due to a combination of polynomial fitting error and finite width fluctuations. We found that these trade off: using a higher-order polynomial fit for $K(c)$ reduces fitting error but then generally requires higher width to achieve the same attenuation of finite-width fluctuations. This is a consequence of the fact that approximating $\E{z \sim \N(0,1)}{|z^k|}$ by random sampling requires exponentially more samples as $k$ increases. We also note that, as shown in Figure \ref{fig:kernel_grid}B, the 4HL ReLU kernel has divergent slope at $c=1$, making low-order polynomial approximations inadequate. In the next section, we provide a solution for these problems, designing a trainable shallow network that mimics a deep ReLU kernel more faithfully and at narrower width.

\subsection{Achieving 4HL Behavior with a 1HL Network}
\label{subsec:exp_mimic}

When choosing an FCN architecture, it is common practice to include different depths in the hyperparameter search. It is of course often the case that the best performance comes from networks with multiple hidden layers. However, our main result suggests that, to the extent that these deep networks operate in the kernel regime, it should be possible to design a 1HL FCN that both trains and generalizes like a desired deep FCN, perhaps using far fewer parameters.

Here we demonstrate this capability by designing a 1HL network that mimics the behavior of a 4HL ReLU network.
This task presents two main difficulties for our Hermite polynomial activations: (1) ReLU NTKs have divergent slope at $c=1$ and are poorly fit by low-order polynomials, and (2) we find that networks with polynomial activations generally diverge upon training.
We circumvent both problems by instead using an activation function with a different parameterization: we begin with an activation function of the variational ansatz
\begin{equation} \label{eqn:phi_mimic}
    \tilde{\phi}(z) = \alpha \, \relu(z - \beta) + \gamma \cos(\delta z + \epsilon) + \zeta z + \eta,
\end{equation}
then numerically optimize $\{\alpha, ..., \eta\}$ such that the NTK of the 1HL network closely matches that of the 4HL ReLU network.
We choose this ansatz because of the observation that ReLU activations give sharply peaked kernels as we require here; in the optimized version, the ReLU term comprises the dominant contribution to $\tilde{\phi}(z)$. The optimized activation function is shown in Figure \ref{fig:mimic_main}A, and its NTK is shown Figure \ref{fig:mimic_main}B.
In this procedure, Theorem \ref{thm:rev} serves as an existence proof of many activation functions achieving the desired kernel, and our numerical procedure allows us to find one with desirable properties.

We next confirm that shallow networks with activation function $\tilde\phi$ indeed mimic the training and generalization behavior of deep ReLU networks. We train width 4096 1HL ReLU, 4HL ReLU, and 1HL $\tilde\phi$ networks on the UCI \texttt{wine-quality-red} task (800 train samples, 799 test samples, 11 features, 6 classes) with full-batch gradient descent, mean-squared-error (MSE) loss, and step size 0.1. As shown in Figures \ref{fig:mimic_main}(C-F), we find that, at all times during training, the train MSE, test MSE, train accuracy, and test accuracy for the engineered shallow net closely track those for the 4HL ReLU network. Like the 4HL ReLU network, the 1HL $\tilde\phi$ network reaches low training error many times faster than the 1HL ReLU network.

Even if their asymptotic performance is the same, shallow FCNs in practice often have the advantage of using far fewer parameters, with the total parameter count scaling as $\O(\textrm{width})$ for shallow models but $\O(\textrm{width}^2)$ for deep models.
It is therefore plausible that our 1HL $\tilde\phi$ network, while matching a 4HL ReLU network in generalization and training speed when both are wide, is in fact superior to a 4HL ReLU network when both have the same total number of parameters.
To test this hypothesis, we train models of many different widths, with appropriate stopping times for each architecture class chosen via cross-validation to maximize validation accuracy. Note that the 1HL ReLU network needed to be trained 10x longer than the other networks for optimal performance.

As shown in Figures \ref{fig:mimic_main}(G,H), at a fixed parameter budget, our engineered shallow network achieves significantly lower training MSE and higher training accuracy than either alternative, suggesting it makes much more efficient use of its parameters in memorizing the training set.
More importantly, as shown in Figures \ref{fig:mimic_main}(I,J), our engineered shallow network consistently outperforms the 4HL ReLU architecture at a fixed parameter budget.
Furthermore, with $> 3 \times 10^4$ parameters, it outperforms the 1HL ReLU network as well.

We repeat this experiment using the \texttt{balance-scale} dataset
(313 train samples, 312 test samples, 4 features, 3 classes), \texttt{breast-cancer-wisc-diag} dataset
(285 train samples, 284 test samples, 30 features, 2 classes),  and report results in Figures \ref{fig:mimic_balance} and \ref{fig:mimic_breast}.
In both cases, we find that the 1HL $\tilde\phi$ network closely mimics the training behavior of the 4HL ReLU net and learned the training data better than the alternatives at fixed parameter budget.
As with the \texttt{wine-quality-red} dataset, on these datasets, the 1HL $\tilde\phi$ network consistently outperforms the 4HL ReLU network at fixed parameter budget and typically outperforms the 1HL ReLU architecture as well.
We also test our procedure on a subset of \texttt{CIFAR-10} \citep{krizhevsky:2009} (1000 train samples, 1000 test samples, 3072 features, 10 classes) and report results in Figure \ref{fig:mimic_cifar10}.
For this high-dimensional dataset, we still see good agreement between the 4HL ReLU and 1HL mimic net, but, due to the dimensionality of the first weight matrix, the 1HL mimic net has lots its advantage in terms of parameter count.

All experiments thus far have normalized train and test data in accordance with Assumption \ref{assm:normalization}.
While this assumption was required by our theory, one might wonder if it is necessary to respect in practice.
To test this, we repeat the above experiments with the \texttt{wine-quality-red} and \texttt{balance-scale} datasets with only normalizing input vectors on average before use.
As reported in Figures \ref{fig:mimic_wqr_unnorm} and \ref{fig:mimic_bs_unnorm}, the results vary somewhat: for the \texttt{wine-quality-red} dataset, the match is still almost perfect in all metrics except test MSE at late times, while for the \texttt{balance-scale} dataset, agreement can be poor depending upon which metric is observed.
We leave the elucidation of what factors affect the quality of agreement for future work, though from NTK theory, one would expect that a larger dataset size would induce greater kernel evolution and thus cause greater divergence.

We stress three important takeaways from these experiments:
\begin{enumerate}
    \item Shallow FCNs can indeed train and generalize like deep FCNs, even at finite width.
    \item Our mimic network was the best-performing architecture for many datasets and parameter budgets, suggesting that our approach (and even our specific $\tilde\phi$) may be practically useful in use cases where parameter efficiency is important.
    \item Most significantly, our kernel engineering paradigm enabled us to design a network architecture in a \textit{principled way}: we began with a clear design goal --- to realize a particular target NTK --- and our final network closely followed expected behavior. We contrast this with the typical black-box trial-and-error method of neural architecture search.\footnote{For example, \citet{ramachandran:2018} also find a high-performing activation function (i.e. Swish), but do so via an expensive exhaustive search that yields little insight into the reasons for high performance.}
\end{enumerate}

It is also noteworthy that, in almost all cases, FCN performance improves essentially monotonically with width (as also seen by \citet{lee:2019-ntk}), suggesting that the limiting kernel regime is optimal and justifying the network kernel as the sole consideration in the design of $\tilde\phi$.

\subsection{Case Study: the Parity Problem}
\label{subsec:exp_parity}

As a second test of our kernel engineering paradigm, we consider the classic \textit{parity problem}. The domain of the parity problem is the boolean cube $\{-1, +1\}^{n_{\textrm{bits}}}$, and the target function is
% \begin{equation}
% f(\xb) = (-1)^{\mathbbm{1}[\xb\textrm{ contains an even \# of 1s}]}
% f(\xb) = (-1)^{\sum_i \frac{\xb_i + 1}{2}}.
% \end{equation}
\begin{equation}
f(\xb) =
\begin{cases}
1 & \xb \textrm{ contains an odd number of +1s}, \\
-1 & \textrm{else}.
\end{cases}
\end{equation}

We take $n_{\textrm{bits}} = 11$ and randomly select half the domain as train data and the other half as test data. Despite its simplicity, the parity problem is notoriously difficult for both FCNs and kernel machines, and despite the fact that a deep ReLU network can easily \textit{represent} the correct solution \citep{hohil:1999, bengio:2007-scaling}, in practice it is not learned during training \citep{bengio:2006-curse, nye:2018}.

The difficulty of the parity problem lies in the fact that points are anticorrelated with their neighbors: flipping any single bit changes the function.
By contrast, most standard FCN kernels have high value when evaluated on nearby points (i.e. $K(c)$ is large when $c$ is nearly but not exactly $1$), so they expect neighboring points to be \textit{correlated}, not anticorrelated.
This can be understood through the lens of the kernel's eigensystem, revealing that, for many FCN kernels, the parity problem is in fact the hardest-to-learn function on the boolean cube \citep{yang:2019-hypercube-spectral-bias, simon:2021-eigenlearning}.

However, we can hope to achieve better performance by engineering a kernel that is better-suited to the problem.
We desire a kernel with a rapid decay as $c$ decreases from $1$.\footnote{Though it might seem we should design a kernel that not only decays but quickly becomes negative as $c$ decreases from $1$, this is in fact impossible for an FCN kernel as it violates the PSD constraint of Theorem \ref{thm:psd}.}
Further noting that the target function has odd symmetry with $n_{\textrm{bits}}=11$, we also want the kernel to be odd.
We would expect that inference with such a kernel predicts correctly on the half of test points that are opposite a training point and predicts close to zero on other test points, giving accuracy of 75\% and MSE of 0.5.

\begin{figure}
  \centering
  \includegraphics[width=8cm]{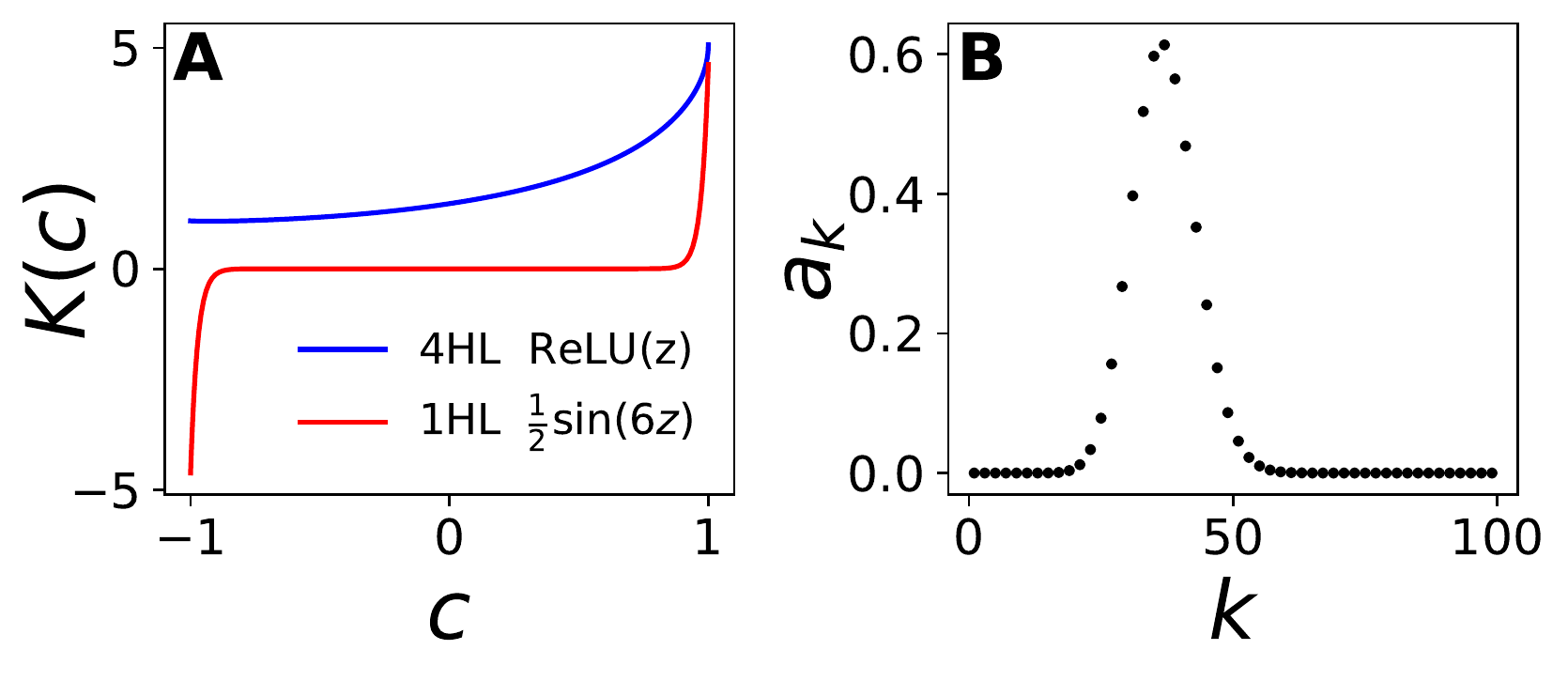}
  \vspace{-3mm}
  \caption{
  \textbf{(A)} NTKs of two networks evaluated on the parity problem. The NTK of the 1HL network with $\frac{1}{2}\sin(6z)$ activations has a rapid decay and odd symmetry, making it well-suited to the parity problem. The NTK of the 1HL network with $10\sin(6z)$ activations (not shown) is identical but $400\times$ greater in magnitude.
  \textbf{(B)} Odd power series coefficients of the NTK of the $1\textrm{HL} \ \frac{1}{2}\sin(6z)$ network. Even coefficients (not shown) are zero.
  }
  \label{fig:parity_kernels}
\end{figure}

We can achieve such a sharp, odd kernel with a sinusoidal activation function. Expanding the kernel of Equation \ref{eqn:sin_kernel} in a power series, we find that
\begin{equation}
    K^{(\text{NTK})}_{\sin(az)}(c) = e^{-a^2} \sum\limits_{\substack{k \ge 1 \\ k \text{ odd}}} \frac{a^{2k}}{k!} (1 + k) c^k,
\end{equation}
which has only odd coefficients that peak at index $k \approx a^2$, which, if $a$ is moderately large, implies the kernel is mostly high-order and thus rapidly decays to zero away from $c=\pm 1$. We choose $a=6$ and test both $\phi(z) = 10 \sin(6z)$ and $\phi(z) = \frac{1}{2} \sin(6z)$. We contrast ReLU and sine kernels in Figure \ref{fig:parity_kernels}.

We trained 4HL ReLU networks and 1HL sine networks via gradient descent with MSE loss.
The results are shown in Table \ref{tab:parity}.
The 4HL ReLU network performs significantly worse than chance\footnote{We define the naive ``chance" predictor as one that always predicts 0 and chooses class labels randomly.}
in terms of both test MSE and test accuracy.
The first 1HL sine network roughly achieved the ``ideal odd kernel" performance we anticipated, which is why we include it in the table.
The second 1HL sine network, however, \textit{genuinely learns the target function}, achieving near-zero MSE and perfect accuracy in every run.
This success surprised us; we hypothesize that it is related to the fact that the parity function can be written as $f(\xb) = \sin(\frac{\pi}{2} \sum_i \xb_i)$.\footnote{The superiority of this shallow architecture is especially noteworthy given that the parity problem was historically used to illustrate the need for deep architectures \citep{bengio:2007-scaling}.}\footnote{The constants in our ultimate sinusoidal activation functions --- $10$, $\frac{1}{2}$, and $6$ --- were chosen after experimentation to best illustrate the two different behaviors highlighted in Table \ref{tab:parity}. Similar choices gave similar results. We leave as an open problem an explanation of why only the smaller prefactor led to perfect generalization. The answer will require ideas beyond the kernel regime.}

\begin{table}[]
\begin{tabular}{cccc}
depth & $\phi(z)$ & test MSE & test accuracy (\%) \\ \hline\hline
4     & $\textrm{ReLU}(z)$     & $2.819 \pm 0.109$ & $29.326 \pm 1.482$ \\
1     & $10 \sin(6z)$          & $0.668 \pm 0.332$ & $82.048 \pm 8.550$ \\
1     & $\frac{1}{2} \sin(6z)$ & $\mathbf{0.021 \pm 0.004}$ & $\mathbf{100 \pm 0}$ \\ \hline
\multicolumn{2}{c}{chance}      & $1$         &         $50$    \\
\multicolumn{2}{c}{ideal odd kernel}      & $.5$         &         75
\end{tabular}
\caption{
\textbf{A kernel-informed choice of activation function dramatically improves performance on the parity problem.}
Test MSE and accuracy are shown for a deep ReLU network and two single-hidden-layer networks with kernel-informed activation functions. Metrics for a naive random predictor and an ideal odd kernel are provided for comparison.
}
\label{tab:parity}
\end{table}

As with the prior experiment, we stress that, though our method does dramatically outperform the FCN baseline on the parity problem, the most important takeaway is not the high performance itself but rather the fact that we achieved it by designing a neural network in a principled way by selecting and reverse-engineering a kernel suited to the task at hand.
Unlike traditional approaches, our method required very little trial-and-error or hyperparameter optimization.

%\subsection{What is the Best Dot-Product Kernel?}
%\label{subsec:exp_kernel_opt}

\section{Conclusions}
\label{sec:conclusions}

Motivated by the need for principled neural architecture design, we have derived a constructive method for achieving any desired positive semidefinite dot-product kernel as an FCN with one hidden layer. In a series of experiments, we verified our construction and showed that this reverse engineering paradigm can indeed enable the design of shallow FCNs with surprising and desirable properties including improved generalization, better parameter efficiency, and deep-FCN behavior with only one hidden layer.

The fact that any FCN kernel can be achieved with just one hidden layer is a surprise with potentially broad implications for deep learning theory.
As with most facets of deep learning, the role and value of depth are not well understood.
Our results suggest that depth may in fact not be important for FCNs.
If this is true, it means that researchers aiming to understand the value of depth should focus on convolutional and other more structured architectures.

Our results open many avenues for future work. On the theoretical side, one might aim to lift the normalization condition on the data or achieve both desired NNGP and NTK kernels simultaneously, likely in a deep network. We also note that the activation functions given by our construction generally have a large number of arbitrary signs, and though all sign choices yield the same limiting kernels, some will outperform others at finite width. Deriving a principle to choose these signs is a potential avenue for future work, and one tool for doing so might be finite-width corrections to the NTK (e.g. \citet{dyer:2019-feynman-diagrams, zavatone:2021}). On the experimental side, there is ample room to apply our method to other tabular data tasks, explore its interactions with regularization and feature learning, and use the large body of work on kernel selection to choose the kernel to reverse engineer \citep{cristianini:2006, cortes:2012, jacot:2020-KARE}.

Though our experiments focus on the NTK, our results also allow the design of a network's NNGP kernel and thus its Bayesian priors. The ability to choose task-appropriate priors instead of defaulting to those of standard architectures would be a valuable tool for the study and practice of Bayesian neural networks (e.g. \citet{wilson:2020-bayesian-DL}), another promising direction for future investigation.

Perhaps the most interesting extension of our work would be the derivation of similar results for convolutional architectures, characterizing their kernel expressivity and finding an analog to our reverse engineering construction. Recent results have shed much light on convolutional kernels' eigenspectra and dependence on hyperparameters \citep{xiao:2021-convnet-eigenspectra, misiakiewicz:2021-convolutional-kernels} so this lofty goal may indeed be within reach. If achieved, we hope the extension of our work to convolutional nets might in time enable the first-principles design of state-of-the-art architectures for the first time.
% in the history of deep learning.

% \bibliography{example_paper}

\section*{Acknowledgements}

The authors would like to thank Chandan Singh, Jascha Sohl-Dickstein, and several reviewers for useful discussions and comments on the manuscript and Jonathan Shewchuk for facilitating our collaboration.
This research was supported in part by the U.S. Army Research Laboratory and the U.S. Army Research Office under contract 
W911NF-20-1-0151.
JS gratefully acknowledges support from the National Science Foundation Graduate Fellow Research Program (NSF-GRFP) under grant DGE 1752814. SA gratefully acknowledges support of the Department of Defense (DoD) through the National Defense Science and Engineering Graduate (NDSEG) Fellowship Program.

\bibliography{ml_refs}
\bibliographystyle{icml2022_modified_maybe}

%%%%%%%%%%%%%%%%%%%%%%%%%%%%%%%%%%%%%%%%%%%%%%%%%%%%%%%%%%%%%%%%%%%%%%%%%%%%%%%
%%%%%%%%%%%%%%%%%%%%%%%%%%%%%%%%%%%%%%%%%%%%%%%%%%%%%%%%%%%%%%%%%%%%%%%%%%%%%%%
% APPENDIX
%%%%%%%%%%%%%%%%%%%%%%%%%%%%%%%%%%%%%%%%%%%%%%%%%%%%%%%%%%%%%%%%%%%%%%%%%%%%%%%
%%%%%%%%%%%%%%%%%%%%%%%%%%%%%%%%%%%%%%%%%%%%%%%%%%%%%%%%%%%%%%%%%%%%%%%%%%%%%%%
\newpage
\appendix
\onecolumn

\section{Additional Figures}
\label{app:additional_figs}
\renewcommand\thefigure{\thesection.\arabic{figure}}

% MIMIC KERNEL FIGURE
\begin{figure}[H]
  \centering
  {\Large \texttt{balance-scale}}
  \includegraphics[width=17cm]{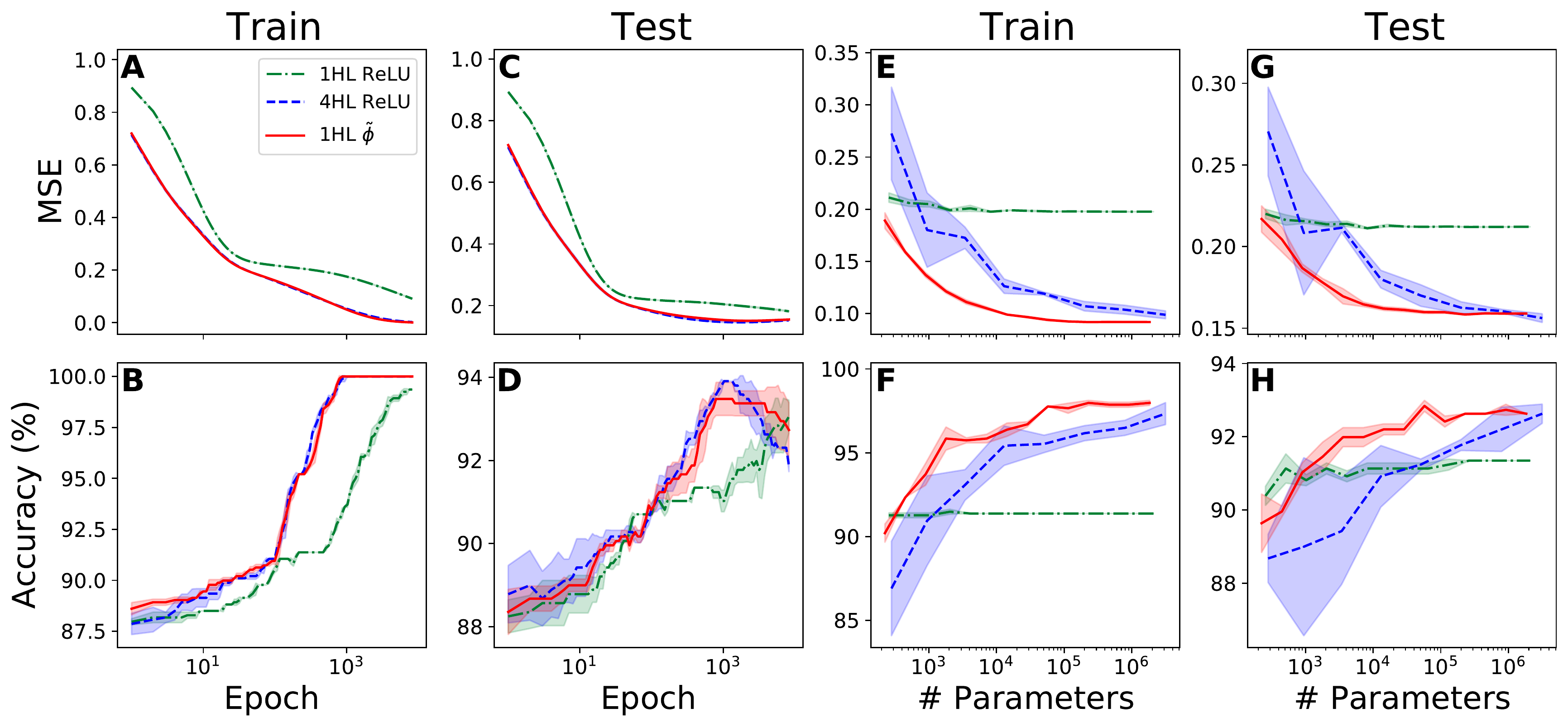}
  \caption{
    \textbf{(A-D)} Train and test MSE and accuracy for the \texttt{balance-scale} task throughout training for finite 1HL ReLU, 4HL ReLU, and 1HL $\tilde\phi$ networks. Metrics for the shallow $\tilde\phi$ network closely match those of the deep ReLU network.
    \textbf{(E-H)}. Final train and test MSE and accuracy for the three architectures with various widths, plotted as a function of the total number of parameters.
  }
  \label{fig:mimic_balance}
\end{figure}

\begin{figure}[H]
  \centering
  {\Large \texttt{breast-cancer-wisc-diag}}
  \includegraphics[width=17cm]{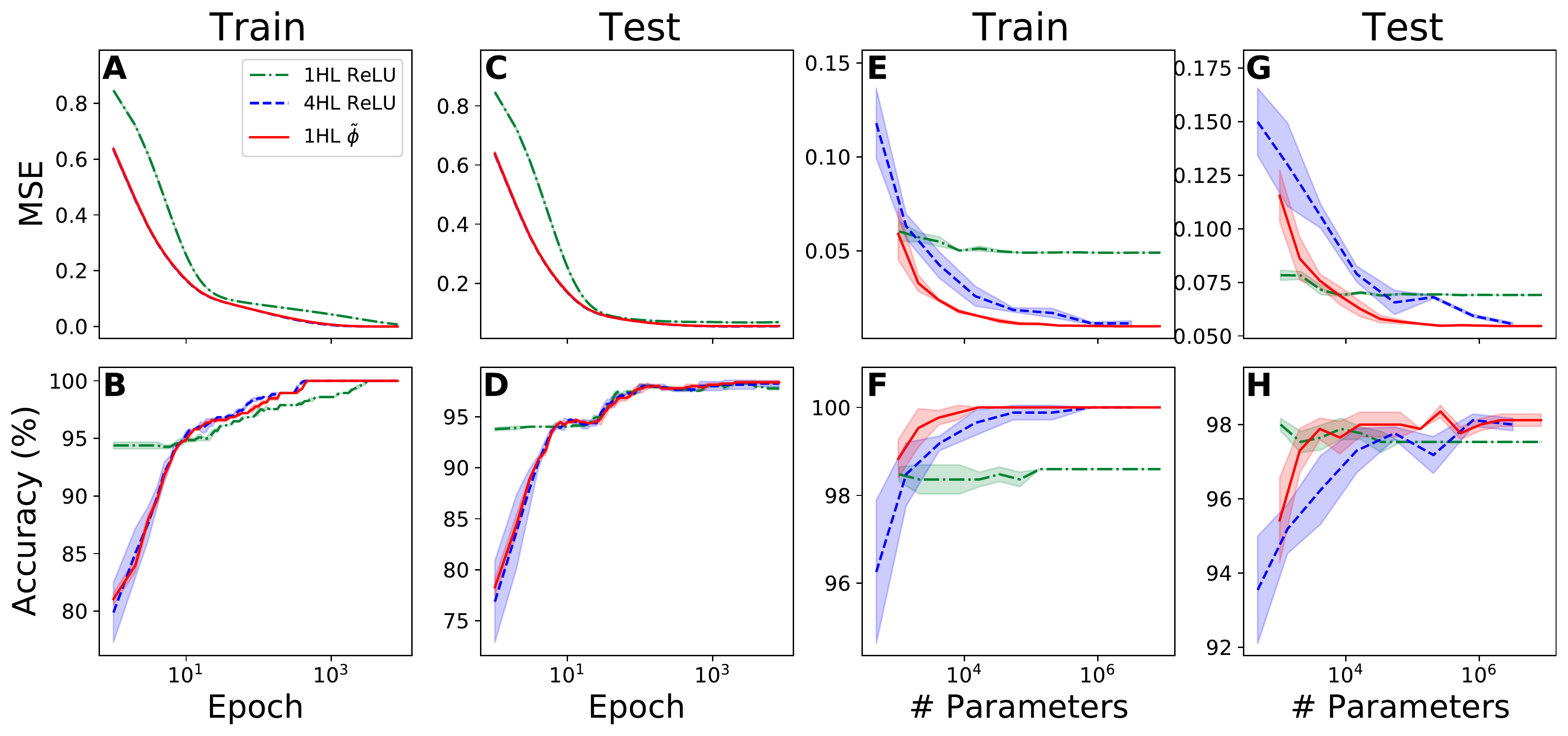}
  \caption{
    \textbf{(A-D)} Train and test MSE and accuracy for the \texttt{breast-cancer-wisc-diag} task throughout training for finite 1HL ReLU, 4HL ReLU, and 1HL $\tilde\phi$ networks. Metrics for the shallow $\tilde\phi$ network closely match those of the deep ReLU network.
    \textbf{(E-H)}. Final train and test MSE and accuracy for the three architectures with various widths, plotted as a function of the total number of parameters.
  }
  \label{fig:mimic_breast}
\end{figure}

\begin{figure}[H]
  \centering
  {\Large \texttt{CIFAR-10}}
  \includegraphics[width=17cm]{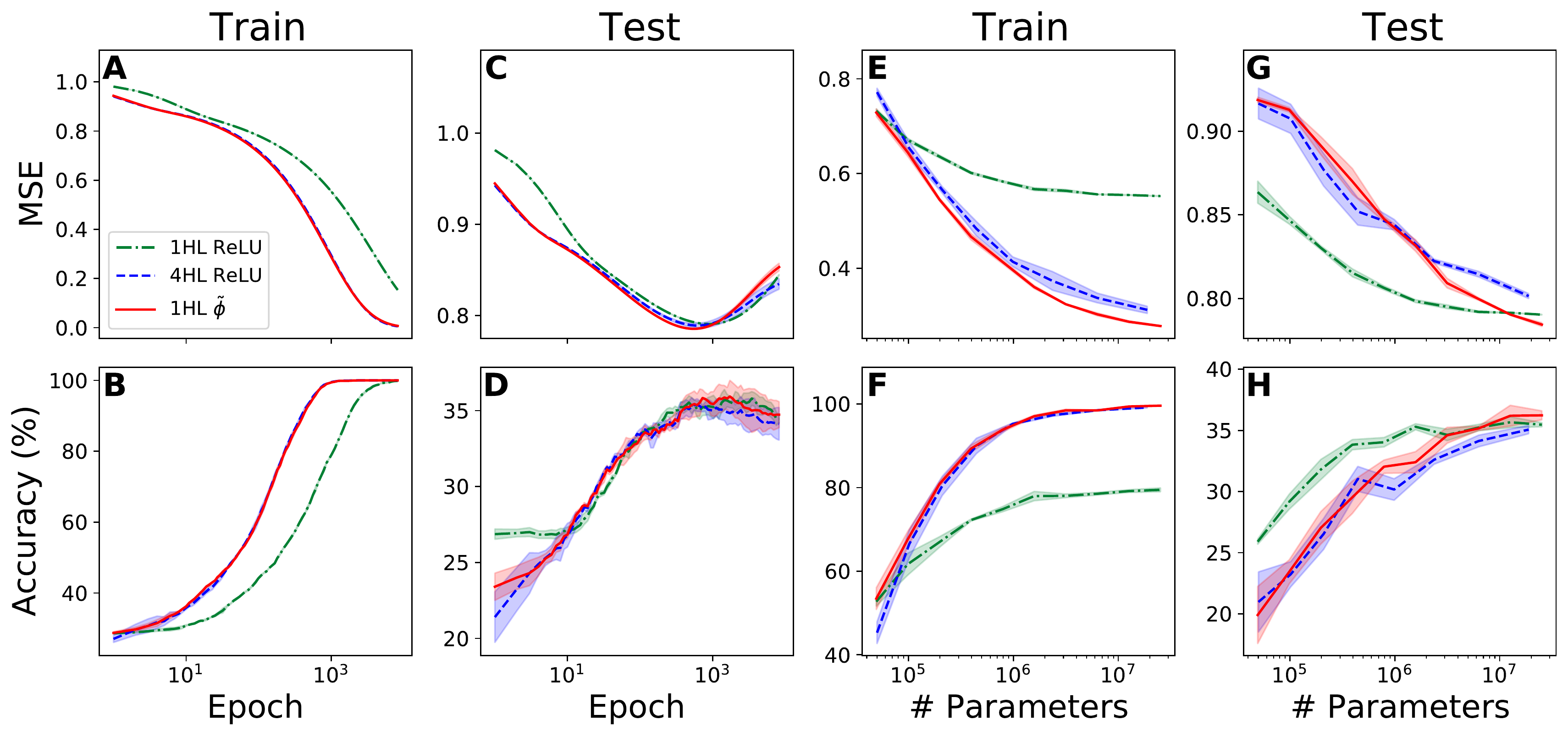}
  \caption{
    \textbf{(A-D)} Train and test MSE and accuracy for a size-1k subset of \texttt{CIFAR-10} throughout training for finite 1HL ReLU, 4HL ReLU, and 1HL $\tilde\phi$ networks. Metrics for the shallow $\tilde\phi$ network closely match those of the deep ReLU network.
    \textbf{(E-H)}. Final train and test MSE and accuracy for the three architectures with various widths, plotted as a function of the total number of parameters. Due to the high input dimension, the 1HL mimic net has little to no advantage over the 4HL ReLU net in terms of parameter efficiency.
  }
  \label{fig:mimic_cifar10}
\end{figure}

\begin{figure}[H]
  \centering
  {\Large \texttt{wine-quality-red (unnormalized)}}
  \includegraphics[width=17cm]{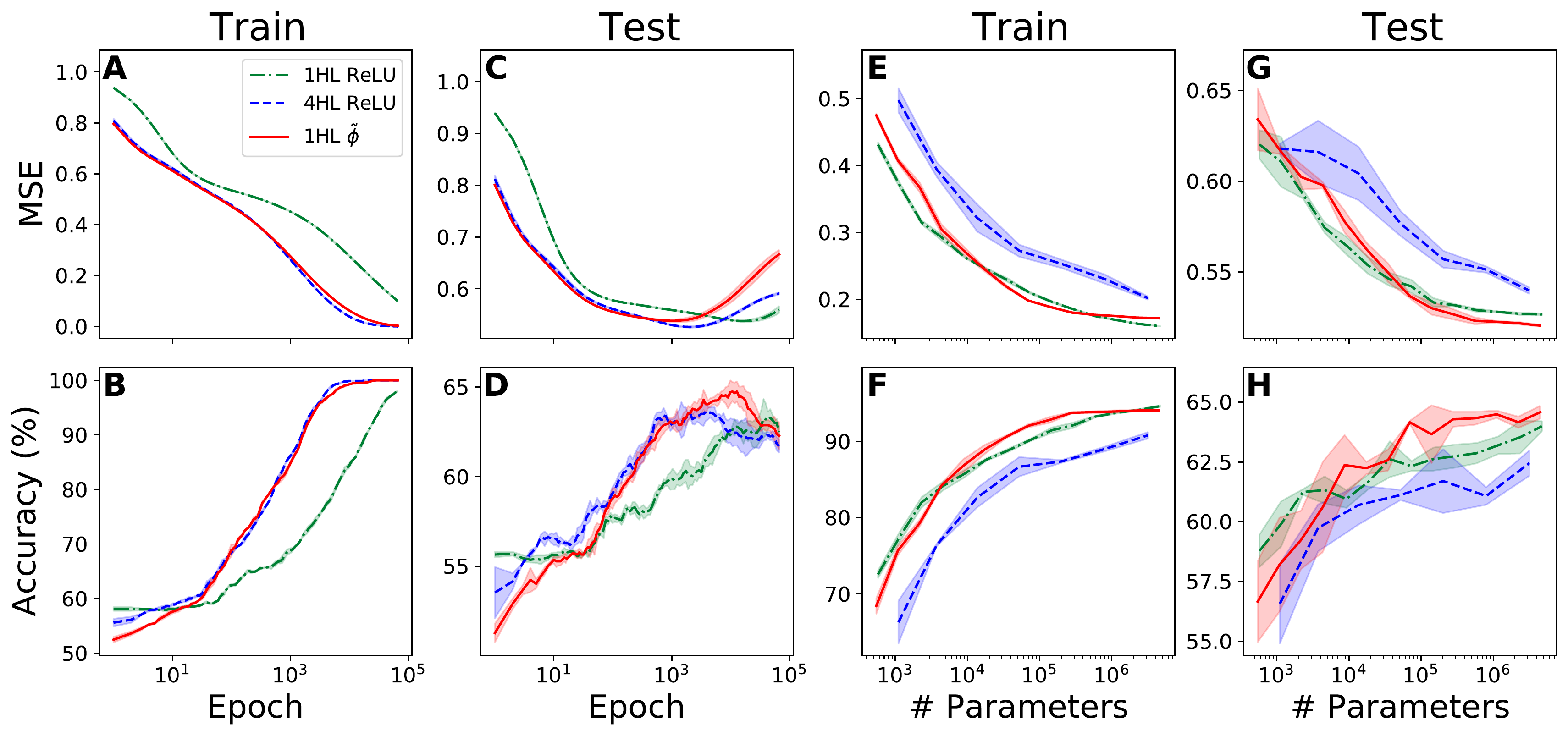}
  \caption{
    \textbf{(A-D)} Train and test MSE and accuracy for the \texttt{wine-quality-red} task \textit{without uniform input normalization}. Metrics for the shallow $\tilde\phi$ network nonetheless still closely match those of the deep ReLU network.
    \textbf{(E-H)}. Final train and test MSE and accuracy for the three architectures with various widths, plotted as a function of the total number of parameters.
  }
  \label{fig:mimic_wqr_unnorm}
\end{figure}

\begin{figure}[H]
  \centering
  {\Large \texttt{balance-scale (unnormalized)}}
  \includegraphics[width=17cm]{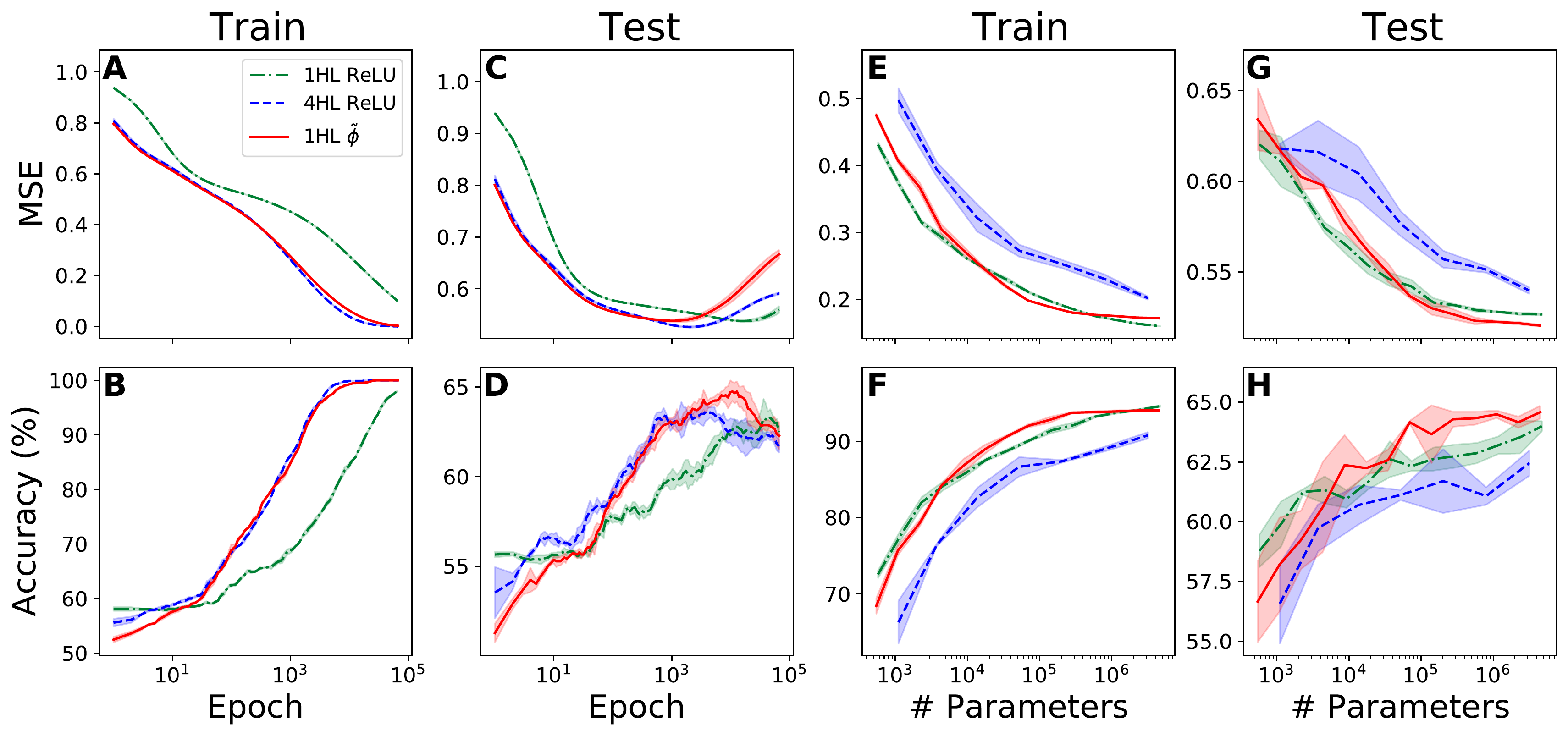}
  \caption{
    \textbf{(A-D)} Train and test MSE and accuracy for the \texttt{balance-scale} task \textit{without uniform input normalization}. For this task, metrics for the shallow $\tilde\phi$ network differ from those of the deep ReLU network.
    \textbf{(E-H)}. Final train and test MSE and accuracy for the three architectures with various widths, plotted as a function of the total number of parameters.
  }
  \label{fig:mimic_bs_unnorm}
\end{figure}

\section{Experimental Details}
\label{app:exp_details}

For all engineered 1HL networks, we use $\sigma_w=1, \sigma_b=0$. For all ReLU and erf networks, we use $\sigma_w=\sqrt{2}, \sigma_b=0.1$ for all layers except the readout layer, for which we use $\sigma_w=1, \sigma_b=0$. We define MSE as $\E{(x,y)}{(f(x)-y)^2}$, without the factor of $\frac{1}{2}$.

Naively, when training an infinitely-wide network, the NTK only describes the \textit{mean} learned function, and the true learned function will include an NNGP-kernel-dependent fluctuation term reflecting the random initialization. However, by \textit{centering} the model --- that is, storing a copy of the parameters at $t=0$ and redefining $\hat{f}_t(x) := \hat{f}_t(x) - \hat{f}_0(x)$ throughout optimization and at test time --- this term becomes zero \citep{lee:2020}, and so we neglect the NNGP kernel in our activation function design and use this trick in our experiments.

\textbf{Arbitrary NTKs with a Single Hidden Layer.}
When fitting a polynomial to $K(c)$, we use a least-squares fit with 10\% of the weight on the point at $c=1$ to encourage better agreement with the 4HL ReLU NTK.

In terms of Hermite polynomials, the activation functions plotted in Figure \ref{fig:kernel_grid} are as follows:

\begin{align}
\phi_A &= 0.837 \, h_0 + 0.271 \, h_1 -0.151 \, h_2 - 0.050 \, h_3 + 0.101 \, h_4 + 0.084 \, h_5, \\
\phi_B &= 1.230 \, h_0 + 0.639 \, h_1 + 0.539 \, h_4 + 0.426 \, h_5, \\
\phi_C &= 0.153 \, h_0 + 0.826 \, h_1 -0.004 \, h_2 -0.005 \, h_3 + 0.219 \, h_4 + 0.577 \, h_5, \\
\phi_D &= 0.001 \, h_0 + 0.010 \, h_1 -0.500 \, h_3 + 0.009 \, h_5, \\
\phi_E &= 0.707 \, h_1 -0.004 \, h_2 -0.003 \, h_3 + 0.447 \, h_4, \\
\phi_F &= 1.001 \, h_1 -0.572 \, h_3 + 0.004 \, h_4 + 0.229 \, h_5, \\
\phi_G &= 1.002 \, h_0 -0.805 \, h_2 + 0.404 \, h_4 + 0.011 \, h_5.
\end{align}

We found that inverting the sign of the $h_2$ and $h_3$ coefficients, while not changing the network's infinite-width NTK, did reduce its finite-width fluctuations and improve the agreement illustrated in Figure \ref{fig:kernel_grid}.
Instead of directly computing the empirical NTK of a single 1HL network with width $2^{18}$, which required too much memory, we equivalently averaged the empirical NTKs of $2^4$ 1HL networks with width $2^{14}$.

\textbf{Achieving 4HL Behavior with a 1HL Network.}

We first optimize the activation function with form given by Equation \ref{eqn:phi_mimic}.
We use the Nelder-Mead algorithm to minimize the sum squared error between the target kernel and the engineered 1HL kernel evaluated on values $c\in[-1,1]$.
The desired kernel is the infinite width 4HL $\relu$ NTK.
One would compare this with the analytical, infinite-width NTK of the 1HL network, but we find that, because our choice of activation function has discontinuous derivative (and is not ReLU), \texttt{neural\_tangents} has difficulty computing its $\tau$-transform, so we instead compare against a randomly-initialized empirical kernel with width $2^{12}$.
To attenuate statistical fluctuations, we average the empirical kernel over 20 initializations. This process yields the activation function
\begin{align}
\tilde{\phi}(z) &= 3.8001 \ \relu(z - 1.0600) -0.0794 \cos(11.8106z + 0.9341) + 0.0968z + 0.9010.
\end{align}

We then use this activation function to train finite width nets and compare the performance of 1HL ReLU, 4HL ReLU, and 1HL $\tilde{\phi}$ networks.
For networks of width $4096$, we train each net for $2^{16}$ epochs, averaging the results over five random initializations.
Using $k=3$-fold cross-validation on the training data, we choose the optimal stopping time for each net and then train networks of varying width, again averaging results over five random initializations.
For the experiments in the main text using the \texttt{wine-quality-red} dataset, we use stopping times of $2000$ for the 4HL ReLU and $\tilde{\phi}$ networks and $20000$ for the 1HL $\tilde{\phi}$ networks.
Experiments on the \texttt{balance-scale} dataset were identical, but used stopping times of $400$ epochs for all networks.
% The fixed-width, variable-time experiments suggested that an even longer runtime may have been beneficial for 1HL $\tilde{\phi}$ networks, but the shorter training time indentified by cross validation may have be a symptom of the small dataset size.
Experiments on the \texttt{breast-cancer-wisc-diag} datasets used stopping times of $700$ epochs for all architectures. Experiments on the \texttt{CIFAR-10} datasets used stopping times of $1000$ epochs for all architectures.

All datasets except \texttt{CIFAR-10} were taken from the UCI repository \citep{dua:2017-UCI-datasets}.

\textbf{Case Study: the Parity Problem.}
We use networks of width 128, trained via full-batch gradient descent with step size $0.1$. We stop when either train MSE is below $10^{-3}$ or after 10k epochs. The 1HL sine networks always stopped due to low train MSE, while the 4HL ReLU networks, which trained much slower, generally timed out. The slow training of ReLU networks here can be understood as a consequence of the presence of very small eigenvalues in their NTK data-data kernel matrix \citep{cao:2019-spectral-bias, yang:2019-hypercube-spectral-bias}. We average performance over 30 trials.

%\textbf{What is the Best Dot-Product Kernel?}

{
}

%%%%%%%%%%%%%%%%%%%%%%%%%%%%%%%%%%%%%%%%%%%%%%%%%%%%%%%%%%%%%%%%%%%%%%%%%%%%%%%
%%%%%%%%%%%%%%%%%%%%%%%%%%%%%%%%%%%%%%%%%%%%%%%%%%%%%%%%%%%%%%%%%%%%%%%%%%%%%%%

\section{Review of Hermite Polynomials} % and Decompositions of Selected Functions}
\label{app:hermite}

% more background on the Hermite polynomials and derive Hermite decompositions of selected activation functions, immediately yielding the NNGP and NTK kernels of the corresponding single-hidden-layer networks. As in the main text, the NNGP results follow from \citet{daniely:2016}, but we include them with the new NTK results for completeness.

As Hermite polynomials are of central importance to our main result, here we provide a more complete introduction.
We note that there are differing conventions for Hermite polynomials; in this work we use the \textit{normalized probabilist's} Hermite polynomials.
The first several such polynomials and the general formula are 
\begin{align*}
    h_0(z) &= 1, \\
    h_1(z) &= z, \\
    h_2(z) &= \frac{1}{\sqrt{2}}\left(z^2-1\right), \\
    h_3(z) &= \frac{1}{\sqrt{3!}}\left(z^3-3z\right), \\
    & \cdots \\
    h_k(z) &= \sum_{\substack{\ell=0 \\ k - \ell \text{\ even}}}^{k} \frac{(-1)^{(k-\ell)/2} \sqrt{k!}}{(k-\ell)!! \ell!} z^\ell, \\
    & \cdots
\end{align*}

In addition to those given in the text, we note the useful identities
\begin{align} \label{eqn:hermite_rr}
    h_k(z) &= \frac{z}{\sqrt{k}} h_{k-1}(z) - \sqrt{\frac{k-1}{i}} h_{k-2}(z), \\
    \label{eqn:hermite_prop_symmetry}
    h_k(-z) &= (-1)^k h_k(z).
\end{align}

Because the Hermite polynomials are orthonormal w.r.t. the standard normal Gaussian metric, we can isolate a particular coefficient of $\phi(z) = \sum_{i=0}^{\infty} b_i h_i(z)$ by computing
\begin{equation} \label{eqn:hermite_isolation}
    b_i = \frac{1}{\sqrt{2\pi}} \int_{-\infty}^{\infty} e^{-z^2 / 2} h_i(z) \phi(z) dz.
\end{equation}
% This computation can be carried out directly for $\phi(z) = e^{az}$ with the use of the aforementioned identities. and the resulting coefficients can be mapped to the corresponding kernel power series using Theorem \ref{thm:rev}, yielding the kernels for a 1HL exponential network. The kernels for 1HL sinusoidal networks are then easy to derive by choosing an imaginary $a$. The ReLU kernels are easiest to derive using known results for $\tau_{\relu}(c)$ \citep{cho:2009} but can also be derived by first decomposing $\phi(z) = \delta(z)$ into the Hermite polynomial basis, finding $\tau_\delta(z)$, and integrating twice using Equation \ref{eqn:tau_tau_dot_identity}.

With the use of Equation \ref{eqn:hermite_prop_double_int}, the Hermite decomposition of an activation function immediately yields its $\tau$-transform (though for special functions there are often faster methods of obtaining it, such as directly computing the expectation in Equation \ref{eqn:tau_transform_def}).
Here we reproduce results from \citet{daniely:2016} and \citet{cho:2009} for the $\tau$-transforms of a few notable functions:
\begin{align}
    \tau_{e^{a z}}(c) &= e^{a^2 (c^2 + 1)}, \\
    \tau_{\sin(a z)}(c) &= e^{-a^2} \sinh(a^2 c), \\
    \tau_{\cos(a z)}(c) &= e^{-a^2} \cosh(a^2 c), \\
    \tau_{\text{ReLU}(z)}(c) &= \frac{1}{2\pi} \left[ (\pi - \cos^{-1}(c)) c + (1 - c^2)^{1/2} \right].
\end{align}

\section{Proofs of Equation \ref{eqn:tau_tau_dot_identity}, Corollary \ref{cor:rev_prop}, and Corollary \ref{cor:impotence_of_depth}}
\label{app:proofs}

\textbf{Proof of Equation \ref{eqn:tau_tau_dot_identity}.} First, we Hermite-expand $\phi$ as $\phi(z) = \sum_{k=0}^\infty b_k h_k(\sigma^{-1} z)$. This yields
\begin{equation}
    \tau_\phi(c; \sigma^2) = \sum_{k=0}^\infty b_k^2 c^k.
\end{equation}
Using Equation \ref{eqn:hermite_prop_derivative} to take the derivatives of $h_k$, we find that $\phi'(z) = \sum_{k=0}^\infty b_k \sigma^{-1} \sqrt{k} h_{k-1}(\sigma^{-1} z)$, and thus
\pushQED{\qed} 
\begin{equation*}
    % \dot{\tau}_\phi(c; \sigma^2) = \sigma^{-2} \sum_{k=0}^\infty b_k^2 k c^{k-1} =
    \tau_{\phi'}(c; \sigma^2) = \sigma^{-2} \sum_{k=0}^\infty b_k^2 k c^{k-1} = \frac{\partial_c}{\sigma^2} \tau_\phi(c; \sigma^2). \qedhere
% \qedsymbol
\end{equation*}
\popQED

A related equation appears in Appendix A of \citet{poole:2016} but is not derived, and this equation for $c=1$ is derived by \citet{daniely:2016}, so we do not claim priority, but we nonetheless hope this proof of the general identity will be useful for the community.

\textbf{Proof of Corollary \ref{cor:rev_prop}}.
\textit{Property (a).} Setting $\sigma_b=0$ in Equations \ref{eqn:1HL_nngp} and \ref{eqn:1HL_ntk} and replacing
% $\dot\tau$
$\tau_{\phi'}$
with $\tau_\phi$ using Equation \ref{eqn:tau_tau_dot_identity} yields that
\begin{align}
    K^{(\text{NNGP})}_\phi(c) &= \sigma_w^2 \tau_\phi(c ; \sigma_w^2), \\
    K^{(\text{NTK})}_\phi(c) &= K^{(\text{NNGP})}(c) + c \partial_c \tau_\phi(c ; \sigma_w^2).
\end{align}
Replacing $\phi$ with $\phi'$ and again using Equation \ref{eqn:tau_tau_dot_identity}, we find that
\begin{align}
    K^{(\text{NNGP})}_{\phi'}(c) &= \partial_c \tau_\phi(c ; \sigma_w^2) = \frac{\partial_c}{\sigma_w^2}K^{(\text{NNGP})}_\phi(c), \\
    K^{(\text{NTK})}_{\phi'}(c) &= K^{(\text{NNGP})}_{\phi'}(c) + \frac{c \partial^2_c}{\sigma_w^2} \tau_\phi(c ; \sigma_w^2) = \frac{\partial_c}{\sigma_w^2}K^{(\text{NTK})}_\phi(c).
\end{align}

\textit{Property (b).} This follows from the fact that $\tau_{\alpha \phi}(c; \sigma^2) = \alpha^2 \tau_{\phi}(c; \sigma^2)$.

\textit{Property (c).} This follows from the fact that $\tau_{\phi(-z)}(c; \sigma^2) = \tau_{\phi(z)}(c; \sigma^2)$.
\pushQED{\qed}
\popQED

\textbf{Proof of Corollary \ref{cor:impotence_of_depth}.} This corollary is true because the kernels of deep FCNs are compositional in their functional form, but certain functions cannot be decomposed into multiple nonlinear PSD polynomials.
To show this, we begin by reproducing the recursion relation for the NNGP kernel (see, e.g., Appendix E of \citet{lee:2019-ntk}).
In terms of the kernel at layer $\ell$, the kernel at layer $\ell + 1$ is
\begin{equation} \label{eqn:nngp_rr}
    K^{(\text{NNGP})}_{\ell+1}(c) = \sigma_w^2 \tau_\phi
    \left( \frac{K^{(\text{NNGP})}_{\ell}(c)}{K^{(\text{NNGP})}_{\ell}(1)} ; K^{(\text{NNGP})}_{\ell}(1) \right) + \sigma_b^2,
\end{equation}
with base condition $K^{(\text{NNGP})}_{0}(c) = \sigma_w^2 c + \sigma_b^2$.
If $\phi$ is nonlinear (i.e. is not an affine function), then the $\tau$-transform in Equation \ref{eqn:nngp_rr} is nonlinear in its first argument, and we have $K^{(\text{NNGP})}_{\ell+1}(c) = f(K^{(\text{NNGP})}_{\ell}(c))$ for some nonlinear PSD function $f$. 

Thus, after two or more nonlinear hidden layers, the NNGP kernel consists of the composition of multiple nonlinear PSD functions. However, certain PSD polynomials cannot be decomposed into multiple nonlinear PSD functions in this way. For example,
\begin{equation}
    K(c) = c^2 + c
\end{equation}
% \begin{align}
%     f(c) &= c^4 + c^2 \\
%     g(c) &= \sqrt{c} \\
%     f(g(c)) &= c^2 + c
% \end{align}
cannot be written as $K(c) = f(g(c))$ for any nonlinear PSD functions $f$ and $g$: to be nonlinear and PSD, they must each have a positive term of at least order two when expressed as a power series, but then their composition must have at least a quartic term, and as $f$ and $g$ are PSD, there can be no negative contribution to cancel this high-order term.\footnote{We note that, while there exist many nonlinear pairs of functions such as $f(c) = c^6 + c^3, \ g(c) = \sqrt[3]{c}$ that yield $K(c) = f(g(c))$, one will always be non-PSD: the flaw with this example is that $\sqrt[3]{c}$ does not have a positive-semidefinite power series expansion and thus cannot be a valid kernel transformation.}
Kernels such as this can only be achieved with exactly one nonlinear hidden layer (though additional hidden layers with \textit{affine} activation functions are fine).

As can be seen from its recursion relations (again see Appendix E of \citet{lee:2019-ntk}), the NTK at any layer takes the form
\begin{equation}
    K^{(\text{NTK})}_{\ell}(c) = K^{(\text{NNGP})}_{\ell}(c) + \textrm{[PSD function]},
\end{equation}
so, if a particular architecture's NNGP kernel must necessarily have extra, unwanted positive terms (such as the inevitable quartic term discussed above), so must the NTK, because the additional PSD contribution cannot cancel any terms.
This completes the proof.

\end{document}